\documentclass[11pt,english,journal]{IEEEtran}
\usepackage[T1]{fontenc}
\usepackage[latin9]{inputenc}
\usepackage{bm}
\usepackage{amsmath}
\usepackage{amssymb}
\usepackage{graphicx}

\makeatletter

\usepackage{amsmath}

\usepackage{enumerate}
\usepackage{color}

\usepackage{times}
\usepackage{mathptmx}
\usepackage{bm}
\usepackage{setspace}

\usepackage{amsthm}
\theoremstyle{plain}

\theoremstyle{plain}

\ifx\proof\undefined\
\newenvironment{proof}[1][\proofname]{\par
   \normalfont\topsep6\p@\@plus6\p@\relax
   \trivlist
  \itemindent\parindent
 \item[\hskip\labelsep
      \scshape
 #1]\ignorespaces
}{%
 \endtrivlist\@endpefalse
}
\providecommand{\proofname}{Proof}
\fi

\makeatother

\usepackage{babel}
\begin{document}
\setstretch{1.091}
\title{Multidimensional Scaling in the Poincar\'e Disk\\~}

\author{Andrej~CVETKOVSKI,~~~~~Mark~CROVELLA
\thanks{Andrej Cvetkovski was with the Department
of Computer Science, Boston University, Boston,
MA, 02215 USA. E-mail: acvetk@gmail.com.}
\thanks{Mark Crovella is with the Department of Computer Science, Boston University, Boston MA, 02215, USA. E-mail: crovella@bu.edu.}
\thanks{This is an electronic pre-print of an article published in NSP Applied Mathematics \& Information Sciences Vol. 10, No. 1, pp. 125-133, 2016, following peer review. The definitive publisher-authenticated version is available as \cite{cv2016}.}
\thanks{The authors acknowledge the National Science Foundation for supporting this work under Grant no. CNS1018266. }
}

\maketitle
\begin{abstract}
Multidimensional scaling (MDS) is a class of projective algorithms traditionally used in \emph{Euclidean space} to produce two- or three-dimensional visualizations of datasets of multidimensional points or point distances. More recently however, several authors have pointed out that for certain datasets, \emph{hyperbolic target space} may provide a better fit than Euclidean space. \\In this paper we develop PD-MDS, a metric MDS algorithm designed specifically for the Poincar\'e disk (PD) model of the hyperbolic plane. Emphasizing the importance of \emph{proceeding from first principles} in spite of the availability of various black box optimizers, our construction is based on an elementary hyperbolic line search and reveals numerous particulars that need to be carefully addressed when implementing this as well as more sophisticated iterative optimization methods in a hyperbolic space model.\end{abstract}

\section{Introduction}

Metric multidimensional scaling (MDS) \cite{cox-cox,BorgGroenen2005}
is a class of algorithms that take as input some or all of the inter-object
distances (\emph{pair dissimilarities}) for $n$ objects and produce
as output a \emph{point configuration} of $n$ points specified by
their coordinates in a chosen $d$-dimensional \emph{target space}. 

The goal is to return the point configuration whose inter-point distances
in the $d$-dimensional space match as closely as possible the original
input distances. Usually, this goal is pursued by minimizing a scalar
badness-of-fit \emph{objective function} defined for an arbitrary
$n$-point configuration in the target space; ideally, the output
of an MDS algorithm should be the configuration that achieves the
global minimum of the objective function. 

If the target space dimension is $2$ or $3$, the output configuration
can be graphically represented, which makes MDS a visualization tool
seeking to preserve the input distances as faithfully as possible,
thus clustering the objects in the target space by similarity. More
generally, for a given dimension $d$, metric multidimensional scaling
can be used to \emph{embed} an\emph{ }input\emph{ }set of dissimilarities
of the original objects into a $d$-dimensional metric space. 

In order to apply MDS, several design decisions must be made. One
first needs to choose a \emph{target metric space }of appropriate
dimension $d$ and a corresponding distance function. An \emph{objective
function }should be chosen so that it provides a suitable measure
of inaccuracy for a given embedding application. If the objective
function is nonlinear but satisfies some mild general conditions (smoothness),
a \emph{numerical optimization }method can be chosen for the implementation.

\subsection{Target space}

The Euclidean plane is the most common choice of a\- target space
for visualization and other applications due to its simplicity and
intuitiveness. Spherical surface can be used, for example, to avoid
the edge effect of a planar representation \cite{cox-sphere}. 

In general, MDS on curved subspaces of Euclidean space can be viewed
as MDS in a higher dimensional Euclidean space constrained to a particular
surface \cite{bentler78-restr-mds,bloxom78-constr-mds}. 

A multidimensional scaling algorithm for fitting distances to constant-curvature
Riemannian spaces is given by \cite{lindman-caelli}. This work uses
the hyperboloid model of the hyperbolic space requiring an $n+1$-dimensional
Euclidean space to represent an $n$-dimensional hyperbolic space,
and is less suitable for visualization purposes. The reader is referred
to \cite{carroll-arabie} or \cite{leeuw-heiser} for a review of
the history of MDS on Riemannian manifolds of constant or nonconstant
curvature.

The use of metric MDS in the hyperbolic plane in the context of interactive
visualization is proposed by \cite{walter}, inspired by the focus
and context hyperbolic tree viewer of \cite{lamping-focus-context}.
The study focuses on the task of embedding higher-dimensional point
sets into 2-dimensional configurations for the purpose of interactive
visualization. It is demonstrated that the PD has capacity to accommodate
lower stress embedding than the Euclidean plane. Several important
pointers to the difficulties one encounters in implementing such algorithms
are given, but a definite specification or implementation is not provided. 

The adequacy of the hyperbolic spaces for embedding of various data
is also studied and confirmed in the contexts of network embedding
for path cost estimation \cite{shavitt} and routing \cite{klein-geohyp,Krioukov:2009:GFS:1639562.1639568,CvetkovskiCrovella:Infocom09,5462131}.

\subsection{Objective function }

A least squares formulation of MDS, to be used in conjunction with
an iterative numerical method for unconstrained optimization is proposed
by \cite{sammon}. The objective function therein (the Sammon stress
criterion) is defined as a normalized sum of the squared differences
between the original dissimilarities and the embedded distances of
the final point configuration. To minimize this function, Sammon proposes
a descent method with step components calculated using the first two
component derivatives of the objective function. 

\cite{walter} adopts Sammon's badness-of-fit measure for hyperbolic
MDS but observes that applying Sammon's iterative procedure in the
Poincar\'{e} disk (PD) using exact derivatives is difficult due to
the complicated symbolic expressions of the second derivative of the
hyperbolic distance function in this model. Subsequently, the Levenberg-Marquardt
least squares method is applied in \cite{walter}, using only first-order
derivatives for the optimization, but the details of applying this
iterative method in the Poincar\'{e} disk are not elaborated. 

\begin{figure}
\noindent \begin{centering}
\includegraphics[width=38mm]{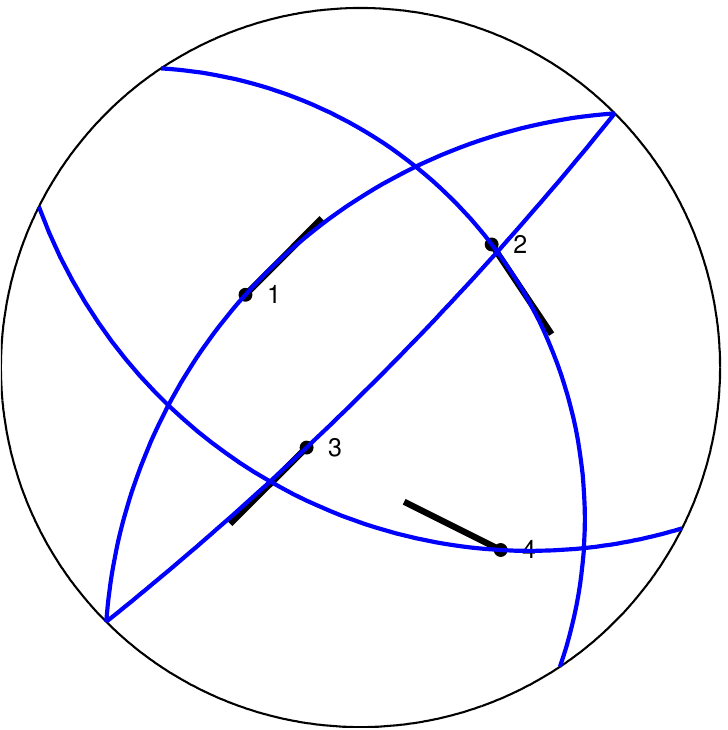}~\includegraphics[width=38mm]{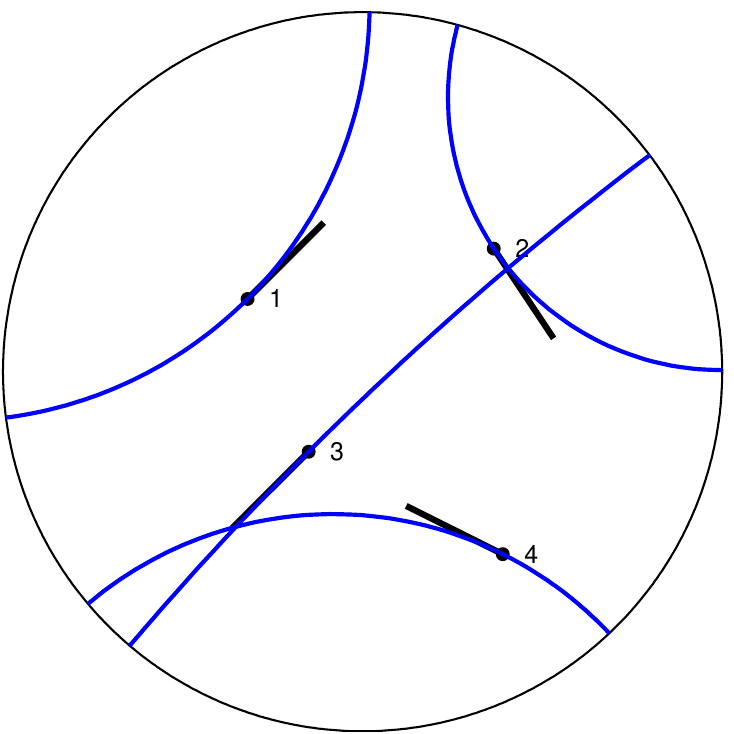}
\par\end{centering}

\noindent \centering{}\caption{\label{fig:Moving-a-configuration-compare}Comparison of the point
trajectories: H-MDS of \cite{walter} (left) vs. PD-MDS hyperbolic
lines (right)}
\end{figure}

The proposed method to convert the seemingly constrained optimization
problem to an unconstrained one by \cite{walter} (Eq.12) ensures
that the moving configuration would stay inside the model during the
optimization. However, this transformation fails to follow the distance
realizing (hyperbolic) lines, or even Euclidean lines. The problem
is illustrated in Fig. \ref{fig:Moving-a-configuration-compare}.
The possibility that the dissimilarity matrix has missing values is
also not addressed in this work, as the dissimilarities are generated
from higher-dimensional points. Input data, however, may also be sparse. 

\subsection{PD-MDS}

In this paper we present PD-MDS, a metric multidimensional scaling
algorithm using the Poincar\'{e} disc (PD) model. For didactic purposes,
we complement our exhibition of PD-MDS with a simple steepest decent
method with line search. We show the details of the steepest descent
along hyperbolic lines in the PD and present a suitable approximate
hyperbolic line search procedure. Based on this development, we show
the particulars of a numerical implementation of PD-MDS.

PD-MDS is applicable in its own right; additionally, its construction
also illustrates some of the specifics that need to be considered
when transferring more sophisticated iterative optimization methods
to the PD or to other hyperbolic models. 

Our numerical experiments indicate that the performance of a steepest
descent method for minimizing a least squares objective on large configurations
in the PD is notably dependent on the line search method used, and
that binary hyperbolic line search provides markedly better convergence
and cost properties for PD-MDS compared to more sophisticated or precise
methods. 

The rest of this paper is organized as follows. Section \ref{sec:2-preliminaries}
consolidates the notation and concepts from hyperbolic geometry that
will be used throughout, and proceeds to develop two of the building
blocks of PD-MDS -- steepest descent in the PD and a corresponding
hyperbolic line search. Section \ref{sec:3-MDS-PD} considers particular
objective functions and gradients and further discusses properties
and applicability of multidimensional scaling in the PD. Section \ref{sec:Numerical-Results}
provides results from the numerical evaluation of the proposed algorithm.
Concluding remarks are given in Section \ref{sec:Concluding-Remarks}.

\section{\noindent \label{sec:2-preliminaries}A descent method for the Poincar\'{e}
disk}

In this section we introduce our notational conventions and establish
some properties of the Poincar\'{e} disk that will be used in what
follows. We then proceed to formally define a Poincar\'{e}-disk specific
descent method and a binary hyperbolic line search, that together
make a simple, yet efficient iterative minimization method for this
model of the hyperbolic plane.

\subsection{Preliminaries}

The \emph{Poincar\'{e} disk model }of the hyperbolic plane is convenient
for our considerations since it has circular symmetry and a closed
form of the inter-point distance formula exists \cite{anderson-hyp}. 

We will be using complex rectangular coordinates to represent the
points of the hyperbolic plane, making the PD model a subset of the
complex plane $\mathbb{C}$: 
\begin{equation}
\mathbb{D}=\left\{ z\in\mathbb{C}\mid\left|z\right|<1\right\} .\label{eq:PDmodel}
\end{equation}

The \emph{hyperbolic distance }between two points $z_{j}$ and $z_{k}$
in $\mathbb{D}$ is given by 
\begin{equation}
d_{\mathbb{D}}\left(z_{j},z_{k}\right)=2\mbox{atanh}\frac{\left|z_{j}-z_{k}\right|}{\left|1-z_{j}\overline{z_{k}}\right|},\label{eq:hypdist}
\end{equation}
 where $\overline{z}$ denotes the complex conjugate.

\emph{M\"{o}bius transformations} are a class of transformations
of the complex plane that preserve generalized circles. The special
M\"{o}bius transformations that take $\mathbb{D}$ to $\mathbb{D}$
and preserve the hyperbolic distance have the form 
\begin{equation}
T\left(z\right)=\frac{az+b}{\overline{b}z+\overline{a}},\,\,\,\, a,b\in\mathbb{C},\,\,\,\,\left|a\right|^{2}-\left|b\right|^{2}\neq0.\label{eq:mobtran}
\end{equation}

Given a point $z_{0}\in\mathbb{D}$ and a direction $\gamma\in\mathbb{C}$
with $\left|\gamma\right|=1$, we can travel a hyperbolic distance
$s\geq0$ along a hyperbolic line starting from $z_{0}$ in the direction
$\gamma$, arriving at the point $z_{0}^{\prime}$.

\textbf{Lemma 1}.~For $z_{0}\in\mathbb{D}$, $\gamma\in\mathbb{C}$
with $\left|\gamma\right|=1$, and $s\geq0$, the point 
\[
z_{0}^{\prime}=\frac{\gamma\tanh\frac{s}{2}+z_{0}}{\overline{z_{0}}\gamma\tanh\frac{s}{2}+1}
\]
 (i) belongs to the hyperbolic ray passing through $z_{0}$ and having
direction $\gamma$ at $z_{0}$, and \\
(ii) $d_{\mathbb{D}}\left(z_{0},z_{0}^{\prime}\right)=s$.

\emph{Proof.} Given a point $z_{0}\in\mathbb{D}$ and a direction
$\gamma\in\mathbb{C}$ with $\left|\gamma\right|=1$, the hyperbolic
ray in $\mathbb{D}$ passing through $z_{0}$ and having direction
$\gamma$ at $z_{0}$ can be parametrized by $r\in\begin{aligned}\left[0,1\right)\end{aligned}
$ as 
\begin{equation}
f\left(r\right)=\frac{r\gamma+z_{0}}{r\gamma\overline{z_{0}}+1}.\label{eq:mob}
\end{equation}

Noting that (\ref{eq:mob}), seen as a function of $z=r\gamma$: 
\[
T\left(z\right)=\frac{z+z_{0}}{z\overline{z_{0}}+1}
\]
is a M\"{o}bius transformation taking $\mathbb{D}$ to $\mathbb{D}$
and preserving hyperbolic distances, we see that 
\[
d_{\mathbb{D}}\left(f\left(r\right),z_{0}\right)=d_{\mathbb{D}}\left(0,r\right)=\ln\frac{1+r}{1-r}
\]
 whence it follows that moving $z_{0}$ along a hyperbolic line in
the direction $\gamma$ by a hyperbolic distance $s=\ln\left(\left(1+r\right)/\left(1-r\right)\right)$
we arrive at the point $z_{0}^{\prime}=f\left(\tanh\frac{s}{2}\right).$
\hfill{}$\square$

Next, we introduce some of the notation that will be used throughout.
\begin{itemize}
\item Let the \emph{point configuration} at iteration $t=1,2,\dots T$ consist
of $n$ points in the Poincar\'{e} disk $\mathbb{D}$ 
\[
z_{j}\left(t\right),\quad j=1\dots n
\]
represented by their rectangular coordinates: 
\[
z_{j}\left(t\right)=y_{j,1}\left(t\right)+iy_{j,2}\left(t\right),\,\, i=\sqrt{-1},\,\, y_{j,1},\, y_{j,2}\in\mathbb{R}
\]
with $\left|z_{j}\left(t\right)\right|<1.$ 
\item We also use vector notation to refer to the point configuration 
\begin{align*}
\mathbf{z}\left(t\right) & =\left[\begin{array}{cccc}
z_{1}\left(t\right) & z_{2}\left(t\right) & \dots & z_{n}\left(t\right)\end{array}\right]^{T}=\mathbf{y}_{1}+i\mathbf{y}_{2}=\\
 & =\left[\begin{array}{cccc}
y_{1,1}\left(t\right) & y_{2,1}\left(t\right) & \dots & y_{n,1}\left(t\right)\end{array}\right]^{T}+\\
 & i\left[\begin{array}{cccc}
y_{1,2}\left(t\right) & y_{2,2}\left(t\right) & \dots & y_{n,2}\left(t\right)\end{array}\right]^{T},
\end{align*}
 where $\left[\cdot\right]^{T}$ in this work indicates the real matrix
transpose (to be distinguished from the complex conjugate transpose.)
\item The \emph{distance matrix} for a given point configuration $\mathbf{z}$
is the real valued symmetric matrix $\mathbf{D\left(\mathbf{z}\right)}=\left[d_{jk}\right]_{n\times n}$
whose entry $d_{jk}$ is the hyperbolic distance between points $z_{j}$
and $z_{k}$ in the configuration $\mathbf{z}$: 
\[
d_{jk}=d_{\mathbb{D}}\left(z_{j},z_{k}\right).
\]
 
\item The \emph{dissimilarity matrix} $\bm{\Delta}=\left[\delta_{jk}\right]_{n\times n}$
is a symmetric, real-valued matrix containing the desired inter-point
distances of the final output configuration (the \emph{dissimilarities}).
The diagonal elements are $\delta_{jj}=0$ and all other entries are
positive real numbers: $\delta_{jk}=\delta_{kj}>0$ for $j\neq k$. 
\item The \emph{indicator matrix} $\mathbf{I}=\left[I_{jk}\right]_{n\times n}$
is a symmetric $0$\nobreakdash-$1$ matrix, used to allow for missing
dissimilarity values. The entries of $\mathbf{I}$ corresponding to
missing values in $\bm{\Delta}$ are set to $0$. All other entries
are set to $1$.
\item The \emph{weight matrix} $\mathbf{W}=\left[w_{jk}\right]_{n\times n}$
is a symmetric, real-valued matrix introduced to enable weighting
of the error terms for individual pairs of points in the objective
function sum. For convenience, $w_{jk}$ corresponding to missing
dissimilarities are set to some finite value, e.g. $1$. 
\item The \emph{objective function} to be minimized is the \emph{embedding
error function} $E_{t}=E_{t}\left(\mathbf{z},\bm{\Delta},\mathbf{W},\mathbf{I}\right)$
that, given the sets of dissimilarities and weights, associates to
a configuration $\mathbf{z}$ an embedding error $E_{t}$. An example
of an error function is the sum of relative squared differences 
\begin{equation}
E_{t}\left(\mathbf{z},\bm{\Delta},\mathbf{W},\mathbf{I}\right)=\sum_{j=1}^{n}\sum_{k=j+1}^{n}w_{jk}I_{jk}\left(\frac{d_{jk}\left(t\right)-\delta_{jk}}{\delta_{jk}}\right)^{2}.\label{eq:stress}
\end{equation}
 The objective function can optionally be normalized per pair by dividing
with the number of summands $\left(n^{2}-n\right)/2$.
\end{itemize}

\subsection{Descent in the Poincar\'{e} disk}

Given a configuration of points $\mathbf{z}$, matrices $\bm{\Delta}$,
$\mathbf{W}$, and $\mathbf{I}$, the distance function $d_{\mathbb{D}}\left(z_{j},z_{k}\right)$,
and an objective function $E\left(\mathbf{z},\bm{\Delta},\mathbf{W},\mathbf{I}\right)$,
define 
\begin{equation}
\mathbf{g}=\nabla E\stackrel{\mbox{def}}{=}\left[\begin{array}{c}
\frac{\partial E}{\partial y_{1,1}}+i\frac{\partial E}{\partial y_{1,2}}\\
\frac{\partial E}{\partial y_{2,1}}+i\frac{\partial E}{\partial y_{2,2}}\\
\vdots\\
\frac{\partial E}{\partial y_{n,1}}+i\frac{\partial E}{\partial y_{n,2}}
\end{array}\right]=\left[\begin{array}{c}
g_{1}\\
g_{2}\\
\vdots\\
g_{n}
\end{array}\right].\label{eq:gE}
\end{equation}

\begin{figure}
\noindent \begin{centering}
\includegraphics[width=50mm]{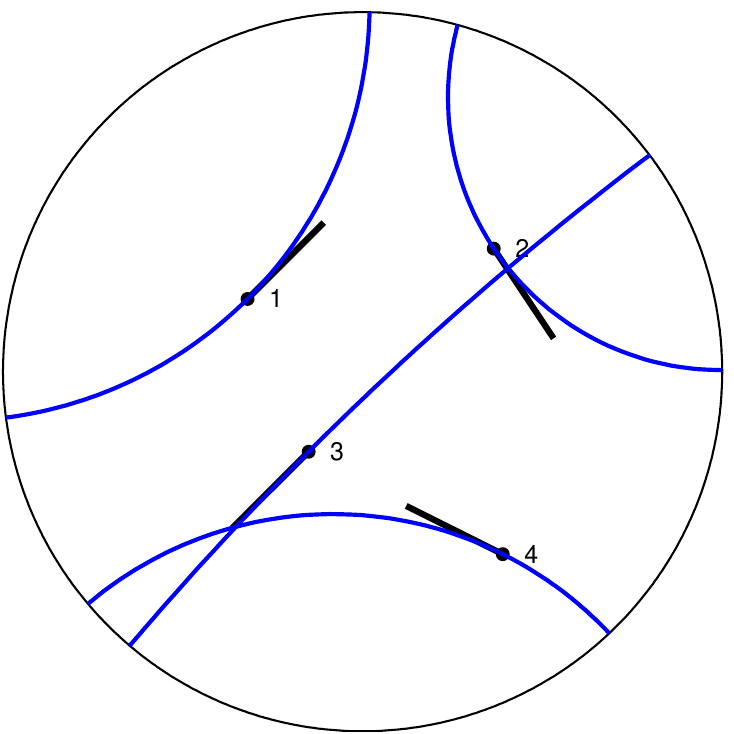}
\par\end{centering}

\noindent \centering{}\caption{\label{fig:Moving-a-configuration}An example of moving a 4-point
configuration in a given (descent) direction along distance realizing
paths of the Poincar\'{e} disk}
\end{figure}

According to Lemma 1, moving the points $z_{1},\dots,z_{n}$ of the
configuration $\mathbf{z}$ along distance realizing paths in the
PD defined respectively by the directions $-g_{1},\dots,-g_{n}$ at\textbf{
$\mathbf{z}$} (Fig. \ref{fig:Moving-a-configuration}) will result
in configuration $\mathbf{z^{\prime}}$ with points 
\begin{equation}
z_{j}^{\prime}=\frac{-rg_{j}+z_{j}}{-rg_{j}\overline{z_{j}}+1}\label{eq:zprime}
\end{equation}
 where $r\geq0$ is the \emph{step-size parameter }which determines
the hyperbolic distances $s_{j}$ traveled by $z_{j}$: 
\begin{equation}
s_{j}=\ln\frac{1+r\left|g_{j}\right|}{1-r\left|g_{j}\right|}.\label{eq:sj}
\end{equation}

The PD model (\ref{eq:PDmodel}) implies the constraints $\left|z_{j}\right|<1$
for the point coordinates. Still, the optimization on the PD can be
viewed as unconstrained by observing that the constraints $\left|z_{j}^{\prime}\right|<1$
will not be violated while moving a configuration $\mathbf{z}$ in
$\mathbb{D}$ if the distances $s_{j}$ traveled by each point are
always kept finite, i.e. 
\begin{equation}
s_{M}=\textrm{max}_{j}s_{j}<\infty.\label{eq:sM}
\end{equation}
Since (\ref{eq:sM}), according to (\ref{eq:sj}), corresponds to
$r\textrm{max}_{j}\left|g_{j}\right|<1$, we have the constraint on
$r$ 
\[
r<\frac{1}{\left\Vert \mathbf{g}\right\Vert _{\infty}}.
\]

When implementing iterative descent minimization methods with line
search in the Poincar\'{e} disk, it is important to specify a hyperbolic
distance window $s_{M}$ along the descent lines where the next configuration
will be sought. In this case the corresponding value of the parameter
$r$ is 
\begin{equation}
r_{M}=\frac{1}{\left\Vert \mathbf{g}\right\Vert _{\infty}}\cdot\tanh\frac{s_{M}}{2}<\frac{1}{\left\Vert \mathbf{g}\right\Vert _{\infty}}.\label{eq:rM}
\end{equation}

Since the Poincar\'{e} disk model is conformal, following the direction
$-\mathbf{g}$ (the opposite of (\ref{eq:gE})) corresponds to the
\emph{steepest descent} optimization method. Moving the point configuration
along hyperbolic lines (distance realizing paths), on the other hand,
ensures that the steepest descent direction is exhausted most efficiently
given the current information about the objective function.

\subsection{A steepest descent algorithm for the PD}

Figure \ref{Flo:algo-hd-mds-1} shows a framework for PD-MDS.

\begin{figure}[h]
{

\hrule height1.2pt width\columnwidth
\vskip 1mm{\normalsize

\textbf{Algorithm }PD-MDS

\hrule height1.2pt width\columnwidth
\vskip 3mm

\textbf{Input data:}

~~an initial configuration $\mathbf{z}\left(1\right)$

~~the dissimilarities $\bm{\Delta}$, weights $\mathbf{W}$, indicators
$\mathbf{I}$

\textbf{Input parameters:} 

~~an objective function $E\left(\mathbf{z},\bm{\Delta},\mathbf{W},\mathbf{I}\right)$

~~the stopping tolerances $\epsilon_{E}$, $\epsilon_{\Delta E}$,
$\epsilon_{\mathbf{g}}$, $\epsilon_{r}$, $T_{M}$

\textbf{Output: }

~~a final point configuration $\mathbf{z}\left(T\right)$

~~a final embedding error $E_{T}$

~

\textbf{Initialize:}

~~~$t\leftarrow1$;~~$s_{M}\leftarrow10$;~~$E_{-1}\leftarrow\infty$;
$\mathbf{z}\leftarrow\mathbf{z}\left(1\right)$;\dotfill{}\{\ref{Flo:algo-hd-mds-1}.1\}

\textbf{Loop:}

~~~~~$E\leftarrow E\left(\mathbf{z},\bm{\Delta},\mathbf{W},\mathbf{I}\right)$;
$\mathbf{g}\leftarrow\nabla E\left(\mathbf{z},\bm{\Delta},\mathbf{W},\mathbf{I}\right)$;\dotfill{}\{\ref{Flo:algo-hd-mds-1}.2\}

~~~~~$r_{M}\leftarrow\frac{1}{\left\Vert \mathbf{g}\right\Vert _{\infty}}\cdot\tanh\frac{s_{M}}{2};$\dotfill{}\{\ref{Flo:algo-hd-mds-1}.3\}

~~~~~\textbf{Break }if 

~~~~~~~~~~~~~$\left.\begin{array}{ll}
 & E<\epsilon_{E}\\
\mbox{\textbf{or}} & E_{-1}-E<\epsilon_{\Delta E}\\
\mbox{\textbf{or}} & \left\Vert \mathbf{g}\right\Vert _{\infty}<\epsilon_{\mathbf{g}}\\
\mbox{\textbf{or}} & r_{M}<\epsilon_{r}\\
\mbox{\textbf{or}} & t>T_{M};
\end{array}\right\} $\dotfill{}\{\ref{Flo:algo-hd-mds-1}.4\}

~~~~~$E_{-1}\leftarrow E;$

~~~~~$r\leftarrow$HypLineSearch$\left(E(\mathbf{z},\bm{\Delta},\mathbf{W},\mathbf{I}),-\mathbf{g},0,r_{M}\right)$;\dotfill{}\{\ref{Flo:algo-hd-mds-1}.5\}

~~~~~$\forall j\in\left\{ 1..n\right\} $, $z_{j}\leftarrow\frac{-rg_{j}+z_{j}}{-rg_{j}\overline{z_{j}}+1}$;\dotfill{}\{\ref{Flo:algo-hd-mds-1}.6\}

~~~~~$t\leftarrow t+1$; 

\textbf{Return }$\mathbf{z}\left(T\right)\leftarrow\mathbf{z}$ and
$E_{T}\leftarrow E\left(\mathbf{z},\bm{\Delta},\mathbf{W},\mathbf{I}\right)$.

}\vskip 1mm
\hrule height1.2pt width\columnwidth

}

\caption{\label{Flo:algo-hd-mds-1}PD-MDS }
\end{figure}

The \emph{input data }of PD-MDS consists of the initial configuration
$\mathbf{z}\left(1\right)$, and the input metric: the dissimilarities
$\bm{\Delta}$ with the associated weights $\mathbf{W}$ and the indicators
of missing  dissimilarities $\mathbf{I}$. 

The \emph{input parameters} are the objective error function $E\left(\mathbf{z},\bm{\Delta},\mathbf{W},\mathbf{I}\right)$
and the stopping tolerances $\epsilon_{E}$, $\epsilon_{\Delta E}$,
$\epsilon_{\mathbf{g}}$, $\epsilon_{r}$, and $T_{M}$. 

The \emph{output }of PD-MDS consists of the final point configuration
$\mathbf{z}\left(T\right)$ and its associated embedding error $E_{T}=E\left(\mathbf{z}\left(T\right),\bm{\Delta},\mathbf{W},\mathbf{I}\right)$.

The \emph{initialization }\{\ref{Flo:algo-hd-mds-1}.1\} sets the
maximum hyperbolic distance $s_{M}$ that can be traveled by any point
of the configuration, and the previous value of the embedding error
$E_{-1}$. 

\emph{Each iteration} starts by determining the gradient of the error
in the current configuration \{\ref{Flo:algo-hd-mds-1}.2\} and the
corresponding window $r_{M}$ \{\ref{Flo:algo-hd-mds-1}.3\} for the
parameter $r$ (Eq. (\ref{eq:rM})). A hyperbolic line search (described
in Sec. \ref{sub:Approximate-Hyperbolic-Line}) is performed \{\ref{Flo:algo-hd-mds-1}.5\}
in the direction of the steepest descent $-\mathbf{g}$ of the embedding
error and the resulting step-size\emph{ }parameter $r$ is used in
\{\ref{Flo:algo-hd-mds-1}.6\} to arrive at the next configuration
as in (\ref{eq:zprime}). 

Several \emph{stopping criteria }are used (line \{\ref{Flo:algo-hd-mds-1}.4\})
to terminate the search. Ideally, the algorithm exits when the embedding
error is close to 0 ($E<\epsilon_{E}$). Termination also occurs in
the cases when the error decreases too slowly ($E_{-1}-E<\epsilon_{\Delta E}$),
or when the gradient or the stepping parameter become too small ($\left\Vert \mathbf{g}\right\Vert _{\infty}<\epsilon_{\mathbf{g}}$,
$r_{M}<\epsilon_{r}$). Finally, $T_{M}$, the maximum allowed number
of iterations, is used as a guard against infinite looping.

The\emph{ line search} subprogram used in \{\ref{Flo:algo-hd-mds-1}.5\}
is described next.

\subsection{\label{sub:Approximate-Hyperbolic-Line}Approximate hyperbolic line
search }

An \emph{exact line search }could be used in line \{\ref{Flo:algo-hd-mds-1}.5\}
(Fig. \ref{Flo:algo-hd-mds-1}) to determine a value for the step
size $r$ such that the corresponding new configuration \{\ref{Flo:algo-hd-mds-1}.6\}
achieves a local minimum of the embedding error along the search path
with tight tolerance: 
\begin{equation}
r\approx\mbox{argmin}_{r\in\left[0,r_{M}\right]}q\left(r\right),\label{eq:idealr}
\end{equation}
 where $q\left(r\right)$ is the embedding error as a function of
$r$. 

However, increasing the precision of this computation is not essential
to the convergence performance since the steepest descent search direction
is only locally optimal. Further, exact line search can fail to converge
to a local minimum even for a second degree polynomial due to finite
machine precision \cite{frandsen}. 

On the other hand, \emph{approximate line search} generally provides
convergence rates comparable to the exact line search while significantly
reducing the computational cost per line search. In fact, the step
calculation used in \cite{sammon} is a ``zero-iteration'' approximate
line search, where the step size is simply guessed based on the first
two derivatives of the error. Conceivably, the simplest inexact step
calculation would guess the step size based only on the directional
gradient at the current configuration. 

Approximate line search procedures aim to reduce the computational
cost of determining the step parameter by posing weaker conditions
on the found solution: Rather than searching for a local or global
minimizer of $q\left(r\right)$ on $\left(0,r_{M}\right]$, a value
is returned by the line search function as satisfactory if it provides
sufficient decrease of the objective function and sufficient progress
toward the solution configuration. A common approach to defining sufficient
decrease is to define the ``roof'' function 
\begin{equation}
\lambda\left(r\right)=q(0)+p\cdot q^{\prime}\left(0\right)\cdot r,\,\,\,\,\,\,\,\,\,\,0<p<1\label{eq:roof}
\end{equation}
which is a line passing through $\left(0,\, q\left(0\right)\right)$
and having a slope which is a fraction of the slope of $q(r)$ at
$r=0$. With this function, we define that sufficient decrease is
provided by all values of $r$ such that

\begin{equation}
q\left(r\right)<\lambda\left(r\right),\,\,\, r\in\left(0,r_{M}\right]\label{eq:accept-decr-cond}
\end{equation}

\begin{figure}
\noindent \begin{centering}
\includegraphics[width=70mm]{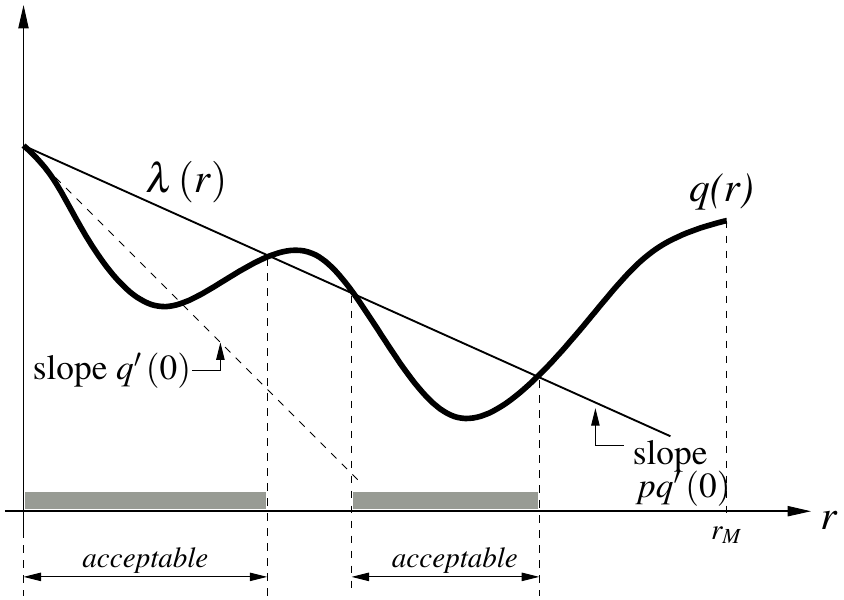}
\par\end{centering}

\noindent \centering{}\caption{\label{fig:linesearch-sketch}Acceptable step lengths for inexact
line search obtained from the sufficient decrease condition.}
\end{figure}

Fig. \ref{fig:linesearch-sketch} shows an example of acceptable step
length segments obtained from the sufficient decrease condition (\ref{eq:accept-decr-cond}). 

To ensure sufficient progress, we adopt a binary search algorithm
motivated by the simple backtracking approach (e.g. \cite{Nocedal1999}).
The details are given in Fig. \ref{Flo:algo-bin-search}. 

\begin{figure}[h]
{

\hrule height1.2pt width\columnwidth
\vskip 1mm{\normalsize

\textbf{Procedure }HypLineSearch

\hrule height1.2pt width\columnwidth
\vskip 3mm

\textbf{Input data:}

~~an initial guess of the step parameter $r_{0}$

~~the maximum step value $r_{M}$

~~the function $q\left(r\right)$

\textbf{Input parameters:} 

~~the slope parameter $p$ for the roof function $\lambda\left(r\right)$;

\textbf{Output: }

~~an acceptable step parameter $r$

~

\textbf{Initialize:}

~~~$r\leftarrow r_{0}$;

\textbf{While }$r<r_{M}$ \textbf{and} $q\left(r\right)<\lambda\left(r\right),$

~~~~~$r\leftarrow2\cdot r$; \dotfill{}\{\ref{Flo:algo-bin-search}.1\}

\textbf{While }$r<r_{M}$ \textbf{or} $q\left(r\right)>\lambda\left(r\right),$

~~~~~$r\leftarrow r/2$;\dotfill{}\{\ref{Flo:algo-bin-search}.2\}

\textbf{Return }$r$.

}\vskip 1mm
\hrule height1.2pt width\columnwidth

}

\caption{\label{Flo:algo-bin-search}Line search procedure for PD-MDS}
\end{figure}

We start the line search with an initial guess $r_{0}$ for the step
size parameter, and in the expansion phase \{\ref{Flo:algo-bin-search}.1\}
we double it until it violates the window $r_{M}$ or the sufficient
decrease condition. In the reduction phase \{\ref{Flo:algo-bin-search}.2\},
we halve $r$ until it finally satisfies both the window requirement
$r<r_{M}$ and the decrease criterion $q\left(r\right)<\lambda\left(r\right)$. 

We observe that, when started at a point with nonzero gradient, the
line search will always return a nonzero value for $r$. Since the
returned acceptable step $r$ is such that the step $2\cdot r$ is
not acceptable, there will be a maximum acceptable point $r_{m}$
from the same acceptable segment as $r$, such that $r\leq r_{m}<2\cdot r$,
whence $r>r_{m}/2$. In other words, the returned value is always
in the upper half of the interval $\left[0,r_{m}\right]$ and we accept
this as sufficient progress toward the solution, thus eliminating
some more computationally demanding progress criteria that would require
calculation of $q^{\prime}(r)$ at points other than $r=0$ or cannot
always return a nonzero $r$ \cite{Nocedal1999,frandsen}.

It remains to show how to calculate the slope of $\lambda\left(r\right)$,
that is $pq^{\prime}\left(0\right)$ (Eq. \ref{eq:roof}). Given a
configuration $\mathbf{z}$ and a direction $-\mathbf{g}=-\nabla E\left(\mathbf{z},\bm{\Delta},\mathbf{W},\mathbf{I}\right)$,
the configuration $\mathbf{z}^{\prime}$ as a function of $r$ (\ref{eq:zprime})
can be conveniently represented as a column-vector function 
\begin{equation}
\mathbf{M}\left(-r\mathbf{g},\mathbf{z}\right)\label{eq:Mfunction}
\end{equation}
whose $j$-th entry is the M\"{o}bius transform 
\[
M_{j}\left(r\right)=\frac{-rg_{j}+z_{j}}{-rg_{j}\overline{z_{j}}+1}.
\]
 The associated embedding error as a function of $r$ is then 
\begin{equation}
q\left(r\right)=E\left(\mathbf{M}\left(-r\mathbf{g},\mathbf{z}\right),\bm{\Delta},\mathbf{W},\mathbf{I}\right),\label{eq:qrhyp}
\end{equation}
and it can be easily shown that its slope is given by
\begin{align*}
q^{\prime}\left(r\right) & =\frac{d}{dr}q\left(r\right)=\\
 & =\left(Re\,\mathbf{M}^{\prime}\left(-r\mathbf{g},\mathbf{z}\right)\right)^{T}Re\,\nabla E\left(\mathbf{M}\left(-r\mathbf{g},\mathbf{z}\right),\bm{\Delta},\mathbf{W},\mathbf{I}\right)\\
 & +\left(Im\,\mathbf{M}^{\prime}\left(-r\mathbf{g},\mathbf{z}\right)\right)^{T}Im\,\nabla E\left(\mathbf{M}\left(-r\mathbf{g},\mathbf{z}\right),\bm{\Delta},\mathbf{W},\mathbf{I}\right)
\end{align*}
where the entries of $\mathbf{M}^{\prime}\left(-r\mathbf{g},\mathbf{z}\right)$
are given by 
\[
M_{j}^{\prime}\left(r\right)=\frac{d}{dr}M_{j}\left(r\right)=g_{j}\frac{\left|z_{j}\right|^{2}-1}{\left(1-rg_{j}\overline{z_{j}}\right)^{2}}.
\]
We thus have a general explicit formula for calculating $q^{\prime}\left(r\right)$
given a configuration $\mathbf{z}$ and the corresponding gradient
$\mathbf{g}$ of $E$ at $\mathbf{z}$. In particular, this formula
can be used to calculate $pq^{\prime}\left(0\right)$, the slope of
$\lambda\left(r\right)$.

\section{\label{sec:3-MDS-PD}Multidimensional scaling in the PD}

\subsection{Objective functions and gradients }

The iterative minimization method presented in Sec. \ref{sec:2-preliminaries}
requires a choice of an embedding error function with continuous first
derivatives. In this work we consider the least squares error function
\begin{equation}
E=c\sum_{j=1}^{n}\sum_{k=j+1}^{n}c_{jk}\left(d_{jk}-a\delta_{jk}\right)^{2}.\label{eq:general_E}
\end{equation}
We note that (\ref{eq:general_E}) is a general form from which several
special embedding error functions can be obtained by substituting
appropriate values of the constants $c$, $c_{jk}$, and $a$. Examples
include: 
\begin{itemize}
\item Absolute Differences Squared (ADS)
\begin{equation}
E=\sum_{j=1}^{n}\sum_{k=j+1}^{n}w_{jk}\left(I_{jk}\left(d_{jk}-a\delta_{jk}\right)\right)^{2}\label{eq:ADS}
\end{equation}
 
\item Relative Differences Squared (RDS)
\begin{equation}
E=\sum_{j=1}^{n}\sum_{k=j+1}^{n}w_{jk}\left(I_{jk}\frac{d_{jk}-a\delta_{jk}}{a\delta_{jk}}\right)^{2}\label{eq:RDS}
\end{equation}

\item Sammon Stress Criterion (SAM)
\begin{equation}
E=\frac{1}{a{\displaystyle \sum_{j=1}^{n}}{\displaystyle \sum_{k=j+1}^{n}}I_{jk}\delta_{jk}}\cdot\sum_{j=1}^{n}\sum_{k=j+1}^{n}w_{jk}\frac{\left(I_{jk}\left(d_{jk}-a\delta_{jk}\right)\right)^{2}}{a\delta_{jk}}\label{eq:SAM}
\end{equation}

\end{itemize}
As the most general case of (\ref{eq:general_E}), individual importance
dependent on the input dissimilarities can be assigned to the pairwise
error terms using the weights terms $w_{jk}$.

PD-MDS also requires calculation of the gradient of the error function.
For a general error function, closed form symbolic derivatives may
or may not exist. In any case, one can resort to approximating the
gradient using finite difference calculations. Numerical approximation
may also have lower computational and implementation costs than the
formal derivatives. However, the use of numerical derivatives can
introduce additional convergence problems due to limited machine precision. 

For the sum (\ref{eq:general_E}), a symbolic derivation of the gradient
of (\ref{eq:general_E}), including both the Euclidean and hyperbolic
cases, can be easily carried out and is omitted here for brevity.
From the obtained result, symbolic derivatives of (\ref{eq:ADS})--(\ref{eq:SAM}),
as well as any other special cases derivable from (\ref{eq:general_E})
can be obtained by substituting appropriate constants.

\subsection{Local vs. global minima}

PD-MDS, being a steepest descent method that terminates at near-zero
progress, can find a \emph{stationary point }of the objective function.
In the least squares case, if the value at the returned solution is
close to zero (that is, $E<\epsilon_{E}$), then the final configuration
can be considered a global minimizer that embeds the input metric
with no error. In all other cases, a single run of PD-MDS cannot distinguish
between local and global points of minimum or between a minimizer
and a stationary point. A common way of getting closer to the global
minimum in MDS is to run the minimization multiple times with different
starting configurations. Expectedly, there will be accumulation of
the results at several values, and the more values are accumulated
at the lowest accumulation point, the better the confidence that the
minimal value represents a global minimum i.e. the least achievable
embedding error. 

Numerous methods that are more likely to find a lower minimum than
the simplest repeated descent methods in a single run have been contemplated
in the numerical optimization literature. However, to guarantee in
general that the global minimizer is found is difficult with any such
method. It may be necessary to resort to running the sophisticated
methods several times as well in order to gain confidence in the final
result. Since these methods are usually computationally more complex
or incorporate a larger number of heuristic parameters, the incurred
computational and implementational costs often offset the benefits
of their sophistication.

\subsection{\label{sec:Curv-matching}Dissimilarity scaling}

The objective functions used in metric \emph{Euclidean }MDS are typically
constructed to be \emph{scale-invariant} in the sense that scaling
the input dissimilarities and the coordinates of the output configuration
with the same constant factor $a$ does not change the embedding error.
This is possible for Euclidean space since the Euclidean distance
function scales by the same constant factor as the point coordinates:
\[
\left(\sum_{s=1}^{L}(a\cdot y_{js}-a\cdot y_{ks})^{2}\right)^{1/2}=a\cdot d_{jk}.
\]
Thus, for example, if $d_{jk}$ is the Euclidean distance, then the
sums (\ref{eq:RDS}) and (\ref{eq:SAM}) are scale-invariant, whereas
(\ref{eq:ADS}) is not. 

However, when $d_{jk}$ is the \emph{hyperbolic }distance function
(\ref{eq:hypdist}), none of the (\ref{eq:ADS})--(\ref{eq:SAM})
are scale-invariant. Therefore, the simplest ADS error function (\ref{eq:ADS})
may be a preferable choice for reducing the computational cost in
the hyperbolic case. 

The lack of scale-invariance of the hyperbolic distance formula (\ref{eq:hypdist})
implies an additional degree of freedom in the optimization of the
embedding error -- the \emph{dissimilarity scaling factor}. In Eqs.
(\ref{eq:general_E})--(\ref{eq:SAM}) this extra degree of freedom
is captured via the parameter $a$ that scales the original entries
of the dissimilarity matrix.

\section{\label{sec:Numerical-Results}Numerical results}

\subsection{A synthetic example}

To illustrate the functioning of PD-MDS, we provide an example random
configuration consisting of seven points in the Poincar\'{e} disk. 

\begin{figure}
\noindent \begin{centering}
\includegraphics[width=60mm]{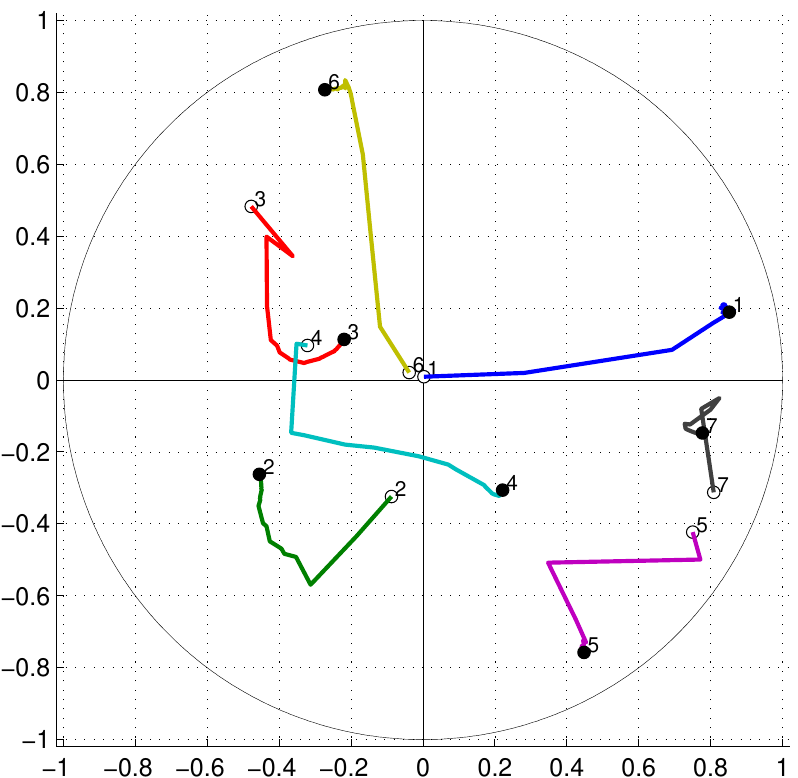} 
\par\end{centering}

\noindent \centering{}\caption{\label{fig:traj-example} The minimization trajectory for a seven
point configuration using PD-MDS. The clear and the solid points are
respectively the initial and the final point configuration.}
\end{figure}

To carry out this experiment, we populate the input dissimilarity
matrix with the hyperbolic inter-point distances and start PD-MDS
from another randomly-generated seven point initial configuration
in the PD. Fig. \ref{fig:traj-example} shows the trajectories traveled
by the points during the minimization. The clear points denote the
initial configuration, whereas the solid ones represent the final
point configuration. 

\begin{figure}
\noindent \begin{centering}
\includegraphics[width=38mm]{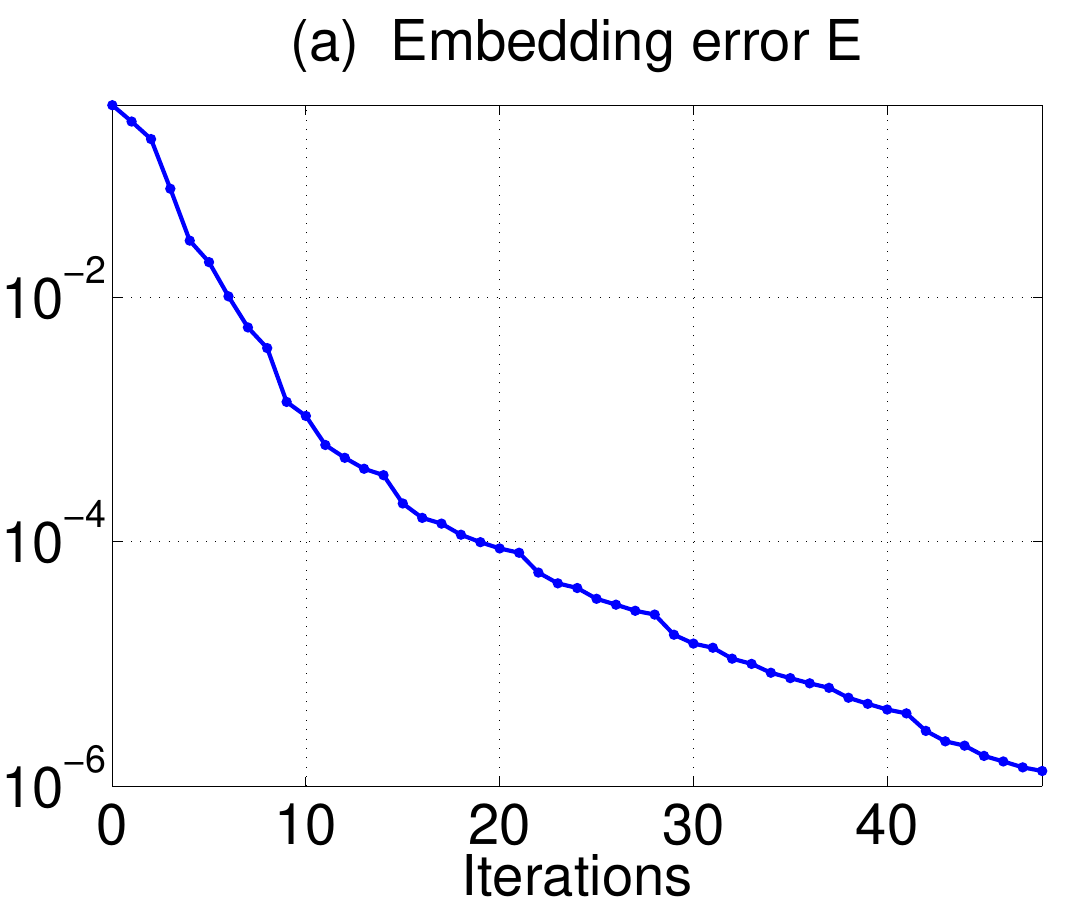}
\includegraphics[width=38mm]{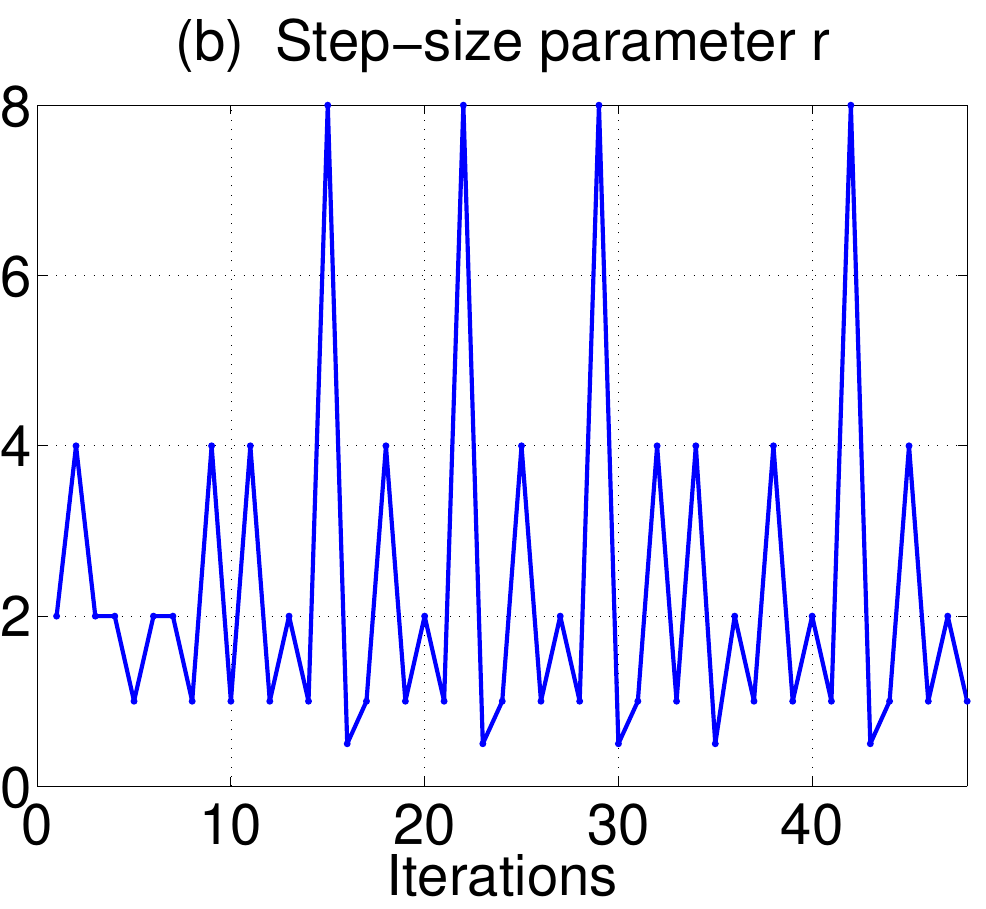}
\par\end{centering}

~

\noindent \begin{centering}
\includegraphics[width=38mm]{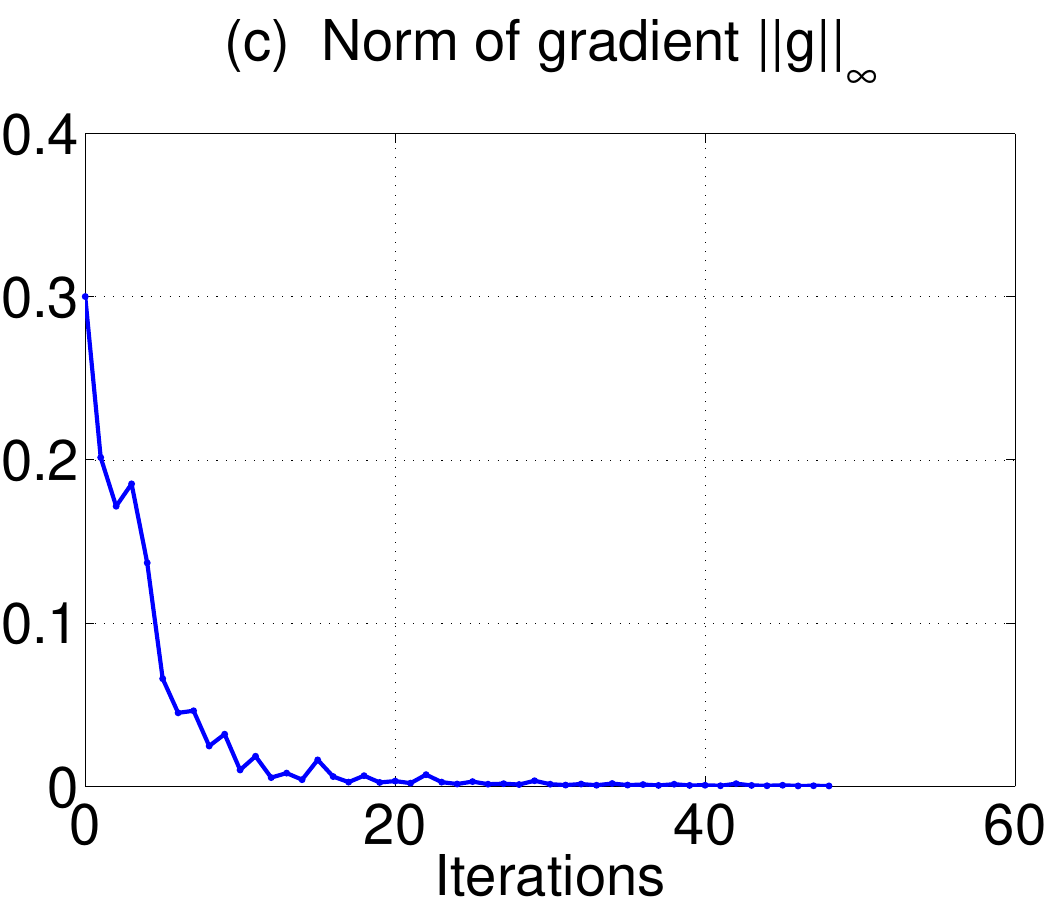}
\includegraphics[width=38mm]{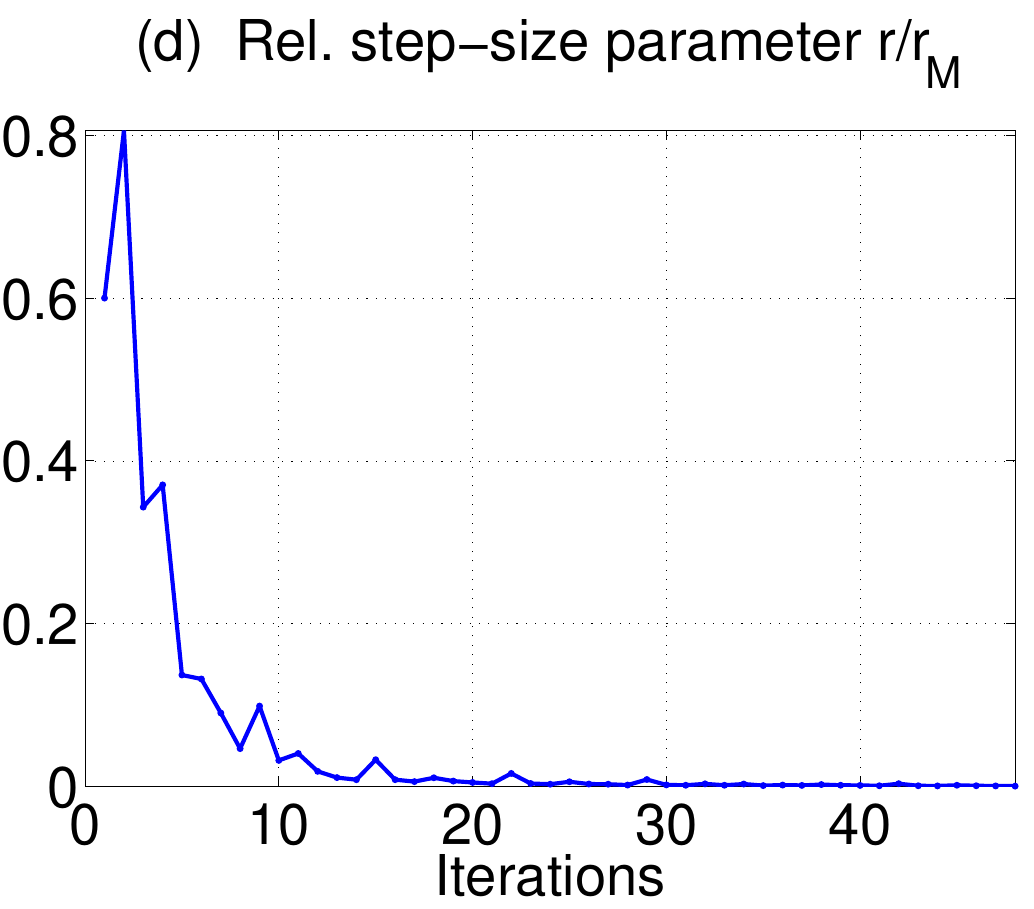}
\par\end{centering}

\noindent \centering{}\caption{\label{fig:traj-example-1} The PD-MDS internal parameters vs. the
iteration number for the seven point example of Fig. \ref{fig:traj-example}:
(a) the embedding error $E$, (b) the step-size parameter $r$, (c)
the norm of the gradient $\left\Vert \mathbf{g}\right\Vert _{\infty}$,
and (d) the step-size parameter relative to the maximum allowed value
$r/r_{M}$. }
\end{figure}

The operation of the PD-MDS algorithm as it iterates over the provided
example configuration is examined in detail in Fig. \ref{fig:traj-example-1}.
The figure shows the PD-MDS internal parameters vs. the iteration
number: In Fig. \ref{fig:traj-example-1}a, the embedding error $E$
monotonically decreases with every iteration; the iterations terminate
at the fulfillment of $E<\epsilon_{E}=10^{-6}$, which means that
likely the output configuration represents the global minimum and
the final inter-point distances match the input dissimilarities very
closely. The step-size parameter $r$ is initialized with a value
of 1 and assumes only values of the form $2^{k}$, for integral $k$
(Fig. \ref{fig:traj-example-1}b). 

The exponential character of the change of $r$ in accord with \{\ref{Flo:algo-bin-search}.1\}
and \{\ref{Flo:algo-bin-search}.2\} (Fig. \ref{Flo:algo-bin-search})
ensures the low computational cost of the line search subprogram. 

The refining of the step size as the current configuration approaches
a local minimum of the error function, on the other hand, is achieved
by the decrease of the gradient norm. This is further illustrated
in Figs. \ref{fig:traj-example-1}c and \ref{fig:traj-example-1}d. 

In our pool of numerical experiments, we produced graphs similar to
those shown in Fig. \ref{fig:traj-example-1} while using two other
line search strategies: (i) exact search and (ii) line search using
an adaptive approximate step-size parameter. Both of these strategies
showed slower convergence compared to the binary hyperbolic line search,
and were of higher computational cost.

\subsection{Scaling of the Iris dataset in the PD}

\begin{figure}
\noindent \begin{centering}
\includegraphics[width=65mm]{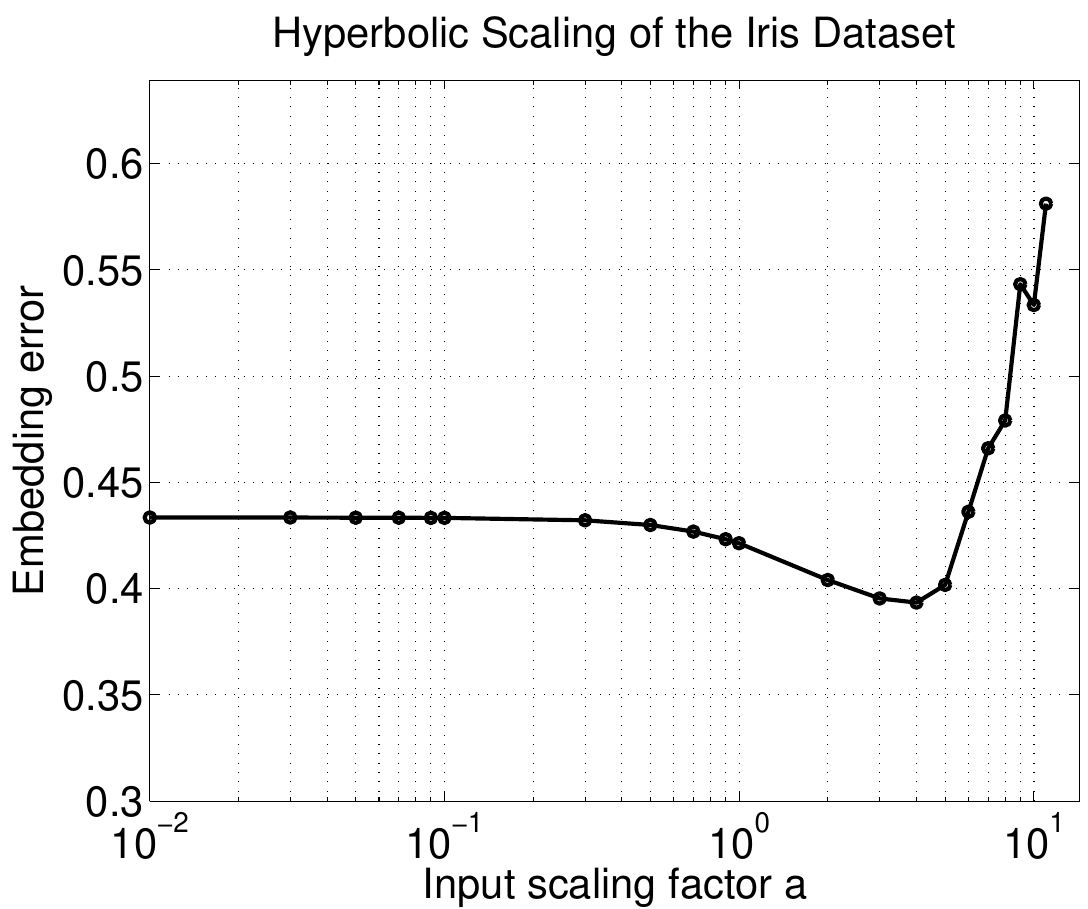}
\par\end{centering}

\noindent \centering{}\caption{\label{fig:iris-scaling}The effect of scaling of the dissimilarities
on the embedding error for the Iris Dataset \cite{anders-iris35}.
The input dissimilarities are the Euclidean distances between pairs
of original points. This PD-MDS result reveals that the Iris dataset
is better suited for embedding to the hyperbolic plane that to the
Euclidean plane. }
\end{figure}

As a first experiment on real-world data, we apply PD-MDS to the Iris
dataset \cite{anders-iris35}. This classical dataset consists of
150 4-dimensional points from which we extract the Euclidean inter-point
distances and use them as input dissimilarities. The embedding error
as a function of the scaling factor $a$ is shown in Fig. \ref{fig:iris-scaling}.
Each value in the diagram is obtained as a minimum embedding error
in a series of 100 replicates starting from randomly chosen initial
configurations. 

Minimal embedding error overall is achieved for $a\approx4$. The
improvement with respect to the 2-dimensional Euclidean case is $10\%$.
Thus, the Iris dataset is an example of dimensionality reduction of
an original higher-dimensional dataset that can be done more successfully
using the PD model. 

\section{\label{sec:Concluding-Remarks}Conclusion}

In this paper, we elaborated the details of PD-MDS, an iterative minimization
method for metric multidimensional scaling of dissimilarity data in
the Poincar\'{e} disk model of the hyperbolic plane. While our exposition
concentrated on a simple steepest descent minimization with approximate
binary hyperbolic line search, we believe that elements of the presented
material will also be useful as a general recipe for transferring
other, more sophisticated iterative methods of unconstrained optimization
to various models of the hyperbolic space.

\section*{Acknowledgement}

The authors acknowledge the National Science Foundation for supporting
this work under Grant no. CNS-1018266.

\end{document}